\documentclass[10pt]{article}

\usepackage[preprint]{tmlr}             %

\usepackage{amsmath,amsfonts,bm}

\def\eqref#1{equation~\ref{#1}}

\def\1{\bm{1}}

\DeclareMathAlphabet{\mathsfit}{\encodingdefault}{\sfdefault}{m}{sl}
\SetMathAlphabet{\mathsfit}{bold}{\encodingdefault}{\sfdefault}{bx}{n}

\usepackage{amsmath,amssymb}
\usepackage{graphicx}
\usepackage{booktabs}
\usepackage{subcaption}
\usepackage{algorithm}
\usepackage{algpseudocode}
\usepackage{xcolor}
\usepackage{url}
\usepackage{hyperref}
\usepackage[capitalise,noabbrev]{cleveref}
\usepackage{placeins}

\title{The Blindfolded Controller: Decision Model-Guided\\ Training for Reinforcement Learning} %

\author{\name Muhammad Ibrahim Khan \email khanm442@coventry.ac.uk \\
        \addr Coventry University, Coventry, UK
        \AND
        \name Aniket Dixit \email dixita4@uni.coventry.ac.uk \\
        \addr Coventry University, Coventry, UK
        \AND
        \name Faizan Ahmed \email ahmedf84@uni.coventry.ac.uk \\
        \addr Coventry University, Coventry, UK
        \AND
        \name James Brusey \email aa3172@coventry.ac.uk \\
        \addr Coventry University, Coventry, UK}

\def\month{MM}
\def\year{YYYY}
\def\openreview{\url{https://openreview.net/forum?id=XXXX}}

\begin{document}

\maketitle

\begin{abstract}
Reinforcement learning (RL) provides a framework for continuous improvement through interaction with an environment. However, RL agents struggle against tasks that offer exploration challenges and sparse rewards, because learning efficient policies from scratch is slow and sample-inefficient. In contrast, decision models (DMs), given a description of the environment and task, can perform well in sparse-reward tasks, but, as they do not learn from reward, they are not able to improve towards the optimal solution. In this paper, we present Decision Model-Guided Training (DMGT), a technique that uses a DM with a description of the environment and task to initialise an RL policy, combining the DM's ability to act without training with RL's ability to improve from reward. Instead of exploring from scratch, the agent reaches the goal and receives reward from its first episodes, and RL then improves its policy beyond what the DM achieves. We demonstrate through experiments on seven tasks that our approach outperforms RL in all seven and DM in five. In addition, we provide comprehensive guidelines, derived from ablation studies, for describing an environment and task to a DM.
\end{abstract}

\section{Introduction}
\label{sec:introduction}

Reinforcement learning learns through interactions with the environment through trial and error. But for tasks with difficult exploration and sparse rewards, an agent starting from scratch can take a lot of timesteps without any reward, so the learning process is slow and data inefficient. For example, in Lunar Lander \citep{towersGymnasiumStandardInterface2024}, during early stages of training PPO crashes because the agent must learn to fire a lander's engine at the beginning of the episode to slow down the falling lander. PPO with default hyperparameters was not able to solve the environment even after a million steps.

A common way to address RL's speed concern is to give the agent a head start. The agent can first imitate demonstrations recorded from human experts \citep{silverMasteringGameGo2016, hesterDeepQlearningDemonstrations2018, rajeswaranLearningComplexDexterous2018}, learn from a teacher policy that already performs the task \citep{schmittKickstartingDeepReinforcement2018, uchenduJumpStartReinforcementLearning2023}, or build on a hand-designed controller \citep{johanninkResidualReinforcementLearning2019, silverResidualPolicyLearning2018}. These approaches show that a good starting policy can substantially reduce the interaction that RL needs. In each case, however, the head start is behaviour that must first be recorded, trained or engineered for the task before RL begins.

A decision model (DM) offers a different perspective, it is a pretrained model that, given a description of the environment, the goal and the possible actions, picks an action without any training on the environment. Hence it is zero-shot: it needs no demonstrations, no reward and no training on the environment, only a description of it.
DM can perform well in sparse-reward and difficult tasks where RL struggles, because a DM's behaviour comes from the description of the environment in natural language and not from trial and error. However a DM does not learn from rewards, therefore it cannot improve from experience, its behaviour is fixed by the model and the description. Given a description of the lander, the pad's location, available actions and the goal, a DM lands the lander on its first attempt, without any training. However, it was only successful for 18 of 20 episodes, and does not have the ability to learn from the unsuccessful episodes.

Therefore RL and DMs have complementary strengths: RL improves through interactions with the environment but starts from scratch, whereas a DM starts with useful behaviour but it cannot improve.
Initialised from a single landing by the DM, PPO succeeded in all 20 test episodes, which PPO from scratch did not, and ends up landing more reliably than the DM itself.

\begin{figure}[t]
  \centering
  \includegraphics[width=0.9\linewidth]{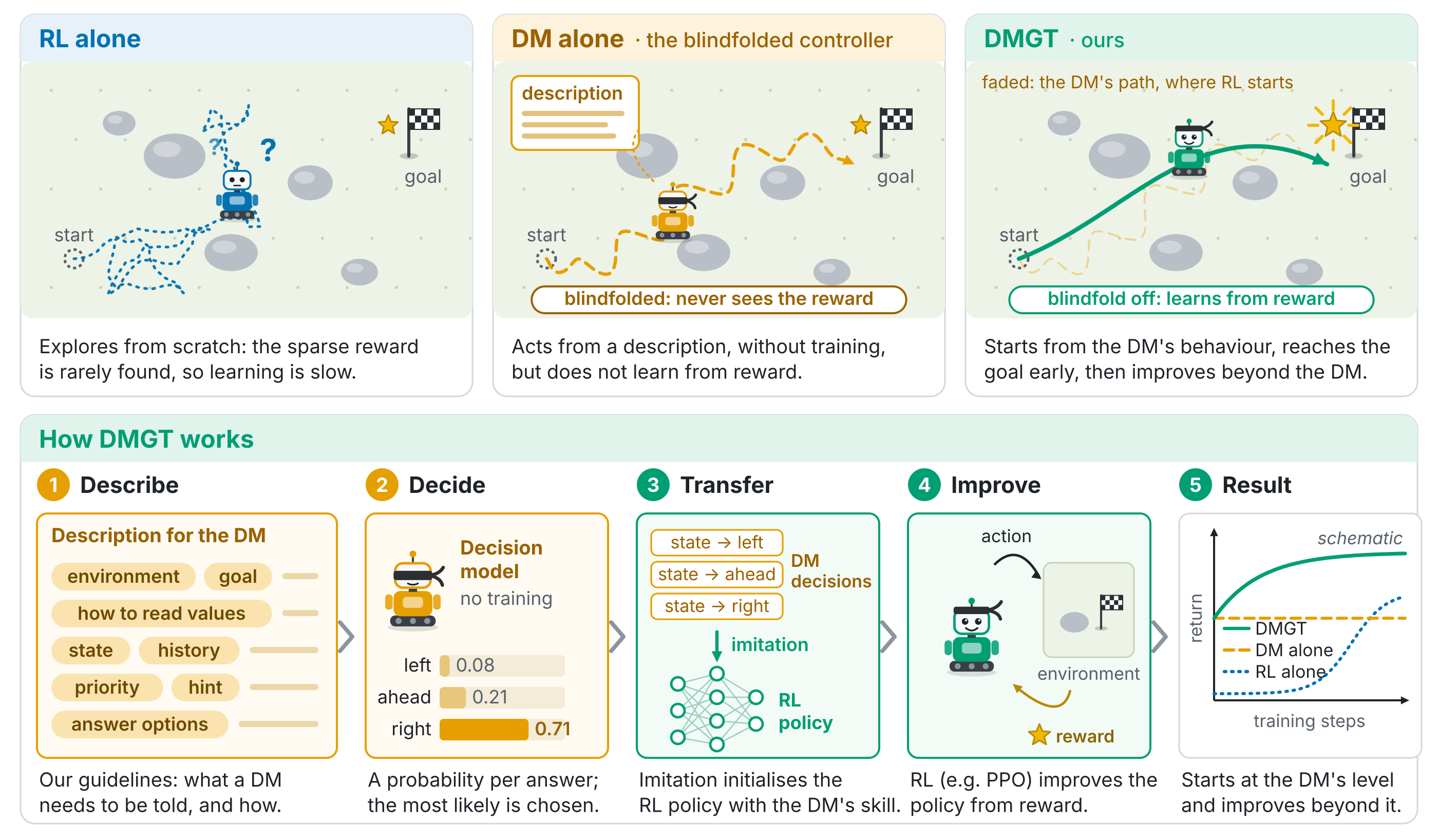}
  \caption{Decision Model-Guided Training (DMGT): combining a decision model (DM) with reinforcement learning (RL). A DM reads a text description of the current situation and returns a probability for each of a fixed set of answers. It needs no training, but the reward is never part of what it reads: it is blindfolded to the reward. \emph{Top:} three agents in the same world, with the reward at the goal. RL alone explores from scratch and rarely finds the reward. The DM alone reaches the goal from its description but cannot learn from the reward. DMGT (ours) starts from the DM's behaviour (the faded path) and then learns from the reward. \emph{Bottom:} how DMGT works. (1)~Each state is described in words, following our guidelines on what a DM needs to be told: the environment, the goal, how to read the values, the state, recent history, a priority (what is most urgent now), a strategy hint and the answer options. (2)~The DM's most likely answer is the action. (3)~A policy network imitates these decisions. (4)~RL (here PPO) improves that network from the reward. (5)~Schematic learning curves: DMGT starts at the DM's level, RL alone from scratch.}
  \label{fig:overview}
\end{figure}

A DM acts based on the description of the task in natural language and the description determines if it succeeds. For example, in Lunar Lander, a DM given only a concise description of the environment lands just as often as with the full description, even though it is told less than a third of the information. What matters is therefore not how much the DM is told, but what information it receives. We study this through an extensive ablation study on seven tasks in six environments, where we remove parts of the description one at a time and then search for the concise and smallest description that still succeeds, and derive guidelines for describing an environment and task to a DM.

We evaluated DMGT on six environments, comparing it with RL alone an with DM alone. In Lunar Lander, PPO initialised from a single landing by the DM reached an average return of 200, after 216k steps, a level that PPO alone was not able to achieve in 1M steps, and its final policy lands in 20 of 20 test episodes, more often than both the DM and PPO. Across all seven tasks, DMGT outperforms PPO alone and converges faster, and it outperforms the DM alone in five of them. Across the six environments, our ablation study shows that the DM cannot act on the raw observations that PPO receives, and an accurate description of each action is essential, and that surface details, such as order of text and number of decimals make no measurable difference.

Our contributions are the following:
\begin{itemize}
    \item We introduce Decision-Model Guided Training (DMGT), a technique that uses a DM, given a description of the environment and task in natural language, to initialise an RL policy through imitation learning from the DM's decisions, RL then further improves the policy.
    \item We study how the description and content given to a DM affect its ability to solve an environment, through ablation studies across seven task in six environments, we derived a guidelines for describing the environment and the task to a DM.
    \item We compare RL alone, the DM alone and DMGT on seven tasks in six environments, and show that DMGT outperforms RL alone in all seven tasks and the DM alone in five tasks.
\end{itemize}

\section{Related Work}
\label{sec:related-work}

\paragraph{Exploration from scratch.}
When rewards are sparse, an RL agent that starts from scratch may act for a long time before it receives any reward. Exploration bonuses based on pseudo-counts \citep{bellemareUnifyingCountBasedExploration2016} or on the error of a learned predictor \citep{pathakCuriositydrivenExplorationSelfsupervised2017, burdaExplorationRandomNetwork2019} reward the agent for reaching novel states, and Go-Explore returns to promising states before exploring further \citep{ecoffetFirstReturnThen2021}. Reward shaping instead adds hand-designed terms that guide the agent towards the goal, and potential-based shaping does so without changing the optimal policy \citep{10.5555/645528.657613}. These methods change how the agent explores or what it is rewarded for, but the agent still starts from a random policy; we start it from the behaviour of a DM instead.

\paragraph{Demonstrations as a head start.}
A common alternative is to begin from demonstrations. AlphaGo first imitated human expert moves and then improved by self-play RL \citep{silverMasteringGameGo2016}. DQfD and DDPGfD add demonstrations to off-policy RL \citep{hesterDeepQlearningDemonstrations2018, vecerikLeveragingDemonstrationsDeep2017}, \citet{nairOvercomingExplorationReinforcement2018} add a behaviour cloning loss, and DAPG pretrains on demonstrations and adds them to the policy gradient \citep{rajeswaranLearningComplexDexterous2018}. PIRLNav and VPT pretrain a policy by behaviour cloning on human demonstrations or on videos with inferred actions, then fine-tune it with RL \citep{ramrakhyaPIRLNavPretrainingImitation2023, bakerVideoPreTrainingVPT2022}. The demonstrations need not be optimal: POfD surpasses the imperfect demonstrations it learns from \citep{kangPolicyOptimizationDemonstrations2018}, and RLPD uses previously collected, possibly suboptimal, data in online RL \citep{ballEfficientOnlineReinforcement2023}. In all of these, the demonstrations or data are recorded on the task beforehand; in our work, they come from a DM given only a description of the environment and task.

\paragraph{Teachers and controllers.}
A head start can also come from a policy that already performs the task. Kickstarting trains a student to match a trained teacher while it learns from reward \citep{schmittKickstartingDeepReinforcement2018}, Jump-Start RL lets a guide policy act for the first part of each episode \citep{uchenduJumpStartReinforcementLearning2023}, and reincarnating RL reuses an earlier agent's policy or data \citep{agarwalReincarnatingReinforcementLearning2022}. Teachers trained with privileged information have been distilled into students that act from their own observations \citep{chenLearningCheating2019, leeLearningQuadrupedalLocomotion2020}, and residual RL learns a correction on top of a hand-designed controller \citep{johanninkResidualReinforcementLearning2019, silverResidualPolicyLearning2018}. Each of these needs a teacher trained on the task or a controller engineered for it.

\paragraph{Imitation learning before RL.}
Behaviour cloning (BC) trains a policy by supervised learning on an expert's states and actions \citep{pomerleauEfficientTrainingArtificial1991}, but its errors compound once the policy drifts from the expert's states, so the bound on its cost grows quadratically with the horizon \citep{rossEfficientReductionsImitation2010}. DAgger reduces this to linear growth by having the expert label the states that the learner itself visits, but it needs an expert that can be queried at any state \citep{rossReductionImitationLearning2011}, which is costly with a human and straightforward with a DM. LOKI similarly runs a few iterations of online imitation before switching to policy-gradient RL \citep{chengFastPolicyLearning2018}. RL fine-tuning can erase what a pretrained policy learned on states that are not visited early in fine-tuning \citep{wolczykFinetuningReinforcementLearning2024}, which VPT and PIRLNav counter with a divergence penalty and a critic warm-up respectively. We use BC and DAgger only to initialise the policy, and fine-tune it with standard PPO.

\paragraph{Descriptions of the task in RL.}
Knowledge of the task can also be given in other forms. Advice-taking learners insert instructions, written in a simple programming language, into the agent's Q-network, which RL then refines \citep{maclinCreatingAdviceTakingReinforcement1996}. Natural language has helped agents through game manuals \citep{branavanLearningWinReading2012}, descriptions of entities \citep{narasimhanGroundingLanguageTransfer2018}, documents describing new dynamics \citep{zhongRTFMGeneralisingNew2020} and instructions turned into shaping rewards \citep{goyalUsingNaturalLanguage2019}; \citet{luketinaSurveyReinforcementLearning2019} survey this line of work. In these methods, the agent learns to use the text by training on the task or on related ones, or the reward model needs annotated examples. A DM, in contrast, acts on a description without any training on the environment. In our work, the DM is used only to initialise the policy through BC or DAgger, after which PPO improves the policy from the environment reward alone, without querying the DM. We also study systematically which parts of a description a DM needs.

\paragraph{Language models as teachers.}
Large language models (LLMs) have also been used as teachers for RL. LLM4Teach distils an LLM's choice among pretrained skills, estimated by sampling its answer several times for the same prompt, into a student RL agent that also learns from reward \citep{zhouLargeLanguageModel2024}. \citet{yanEfficientReinforcementLearning2025} turn several free-form LLM answers per state into actions with a hand-written rule and use them as a prior over actions, while instructRL asks an LLM, one action at a time, whether to take it under a human instruction and regularises RL towards the resulting prior \citep{huLanguageInstructedReinforcement2023}. A DM is a different kind of model altogether: it is built to decide rather than to write, and works like a classifier whose classes are the options given in its query, returning a probability for each in a single pass. Its answer is therefore always one of those options, there is no output to parse and nothing that can fail to parse, and one call yields the whole distribution. This makes a DM far faster and cheaper to run than a teacher that must generate text, be sampled repeatedly or be asked once per action.
\section{Background}
\label{sec:background}

\subsection{Reinforcement Learning}
\label{sec:background-rl}
In reinforcement learning (RL), an agent learns to act in an environment by interacting with the environment and receiving rewards \citep{suttonReinforcementLearningIntroduction2018}.  We model the environment as a Markov decision process (MDP) $(\mathcal{S}, \mathcal{A}, P, r, \mu, \gamma)$. An episode starts in a state $s_0 \sim \mu$, and at each step $t$, the agent receives an observation $o_t$ of the current
state $s_t \in \mathcal{S}$ and takes an action $a_t \sim \pi(\cdot \mid o_t)$ from a discrete or continuous action
space $\mathcal{A}$. It then receives a reward $r_t = r(s_t, a_t)$, and the environment moves to a new state
$s_{t+1} \sim P(\cdot \mid s_t, a_t)$. RL seeks a policy $\pi$ that maximises the expected return
$J(\pi) = \mathbb{E}_\pi\big[\sum_{t=0}^{H-1} \gamma^t r_t\big]$ over an episode of $H$ steps, where the discount
factor $\gamma \in [0, 1)$ weights near rewards more than distant ones.

We use Proximal Policy Optimization (PPO) \citep{schulmanProximalPolicyOptimization2017}, an on-policy actor-critic
method with a policy (the actor) $\pi_\theta$ and a value function (the critic) $V_\psi(o)$ that estimates the expected
return from observation $o$. PPO alternates between collecting a batch of steps with the current policy and updating
$\theta$ for several epochs on that batch to maximise
\begin{equation}
  L^{\mathrm{CLIP}}(\theta) = \mathbb{E}_t\Big[\min\big(\rho_t(\theta)\hat{A}_t,\;
    \operatorname{clip}\big(\rho_t(\theta), 1-\epsilon, 1+\epsilon\big)\hat{A}_t\big)\Big],
  \qquad
  \rho_t(\theta) = \frac{\pi_\theta(a_t \mid o_t)}{\pi_{\theta_{\mathrm{old}}}(a_t \mid o_t)},
  \label{eq:ppo}
\end{equation}
where $\theta_{\mathrm{old}}$ are the parameters that collected the batch and $\epsilon$ limits how far one update can
move the policy. The advantages $\hat{A}_t$ are computed from the rewards and $V_\psi$ by generalized advantage
estimation \citep{schulmanHighDimensionalContinuousControl2016}, and $V_\psi$ is trained alongside $\pi_\theta$ to
predict the returns.

\subsection{Decision Models}
\label{sec:background-decision-models}
A decision model (DM) is a pretrained model that chooses among given options. Examples include Jev \citep{typesafeaiJevModels2026}, the Decision~3.0 models d3-flash and d3-turbo \citep{vllmsemanticrouterteamD3flashMultimodalFoundation2026,vllmsemanticrouterteamD3turboMultimodalDecision2026}, Clef \citep{cloudflareClef2026}, Perplexity's Decider \citep{perplexityPplxdeciderv122026} and Laya \citep{convaiinnovationsLaya2026}. To ask a decision model to make a decision, we send it a query $x$ written in natural language. The query has three parts: a description of the situation, a question saying what is to be decided, and a finite set of answer options $\mathcal{Y}$, each with a very concise description of what it means. The DM returns a probability $M(y \mid x)$ for each option $y \in \mathcal{Y}$, and its choice is always the most probable option. We only query DMs, they are never trained or fine-tuned on the environment therefore they are zero-shot.

A DM can therefore perform well in an environment it has not been trained on. At every step, the query describes the environment, the task and the current state, and each action. Applying the chosen option's action at every step defines the DM policy $\pi_M$, whose action at step $t$ we write $a^M_t$. Unlike an RL policy, a DM policy does not improve with experience.

\subsection{Imitation Learning}
\label{sec:background-imitation}
Imitation learning trains a learner policy to act like an expert. Behaviour cloning (BC) \citep{pomerleauEfficientTrainingArtificial1991} records the expert's observations and actions in a dataset $\mathcal{D}$ and trains the learner by supervised learning, typically by minimising
$-\sum_{(o, a) \in \mathcal{D}} \log \pi(a \mid o)$. Because BC trains only on states the expert visits, a learner
that makes a mistake can reach states the expert never visited, and its errors can compound over the episode
\citep{rossEfficientReductionsImitation2010}. DAgger \citep{rossReductionImitationLearning2011} reduces this by also
training on the states the learner reaches: it repeatedly runs the current learner, has the expert label the
observations it visited, adds them to $\mathcal{D}$ and trains a new learner on all of $\mathcal{D}$. This requires an
expert that can be asked for its action in any state, not only one that can record demonstrations.
\section{Decision Model-Guided Training}
\label{sec:method}

In Decision Model-Guided Training (DMGT), an RL agent do not start from scratch, it begins with the behaviour of a DM and then improves on it. DMGT has three stages, as illustrated in the lower half of \cref{fig:overview}. First, we describe the environment and the task in natural language to the DM, which then decides an action at every step without seeing the environment or the reward (\cref{sec:querying}). Second, we transfer this behaviour to the RL policy by imitation learning. The DM's choices become labels for a network that sees only the environment's observations (\cref{sec:learning-from-dm}). Third, PPO then improves this policy by repeated interactions with the environment and the reward. The DM is used only before RL begins: PPO, and the policy it returns, act without it.

\begin{algorithm}[t]
  \caption{Decision Model-Guided Training. $\phi$ turns the history $h_t$ (observations and actions so far) into a
    query: the description of the environment and task, the current state, and one answer option per action.
    $a^M_t = \arg\max_{a \in \mathcal{A}} M(a \mid \phi(h_t))$ is the decision model's action, and
    $\textsc{Fit}(\mathcal{D}) = \arg\min_{\pi} \sum_{(o, a) \in \mathcal{D}} -\log \pi(a \mid o)$ trains a new
    student on $\mathcal{D}$.}
  \label{alg:method}
  \begin{algorithmic}[1]
    \Require environment $\mathcal{E}$ with actions $\mathcal{A}$; decision model $M$; description function $\phi$;
      numbers of demonstration episodes $N$, DAgger iterations $K$, episodes per iteration $E$ and PPO steps $T$
    \State Run $M$ in $\mathcal{E}$ for $N$ episodes, taking $a^M_t$; store $(o_t, a^M_t)$ for every step in
      $\mathcal{D}$
    \State $\pi_0 \gets \textsc{Fit}(\mathcal{D})$ \Comment{behaviour cloning}
    \For{$k = 1, \dots, K$} \Comment{DAgger; $K = 0$ skips it}
      \State Run $\pi_{k-1}$ greedily in $\mathcal{E}$ for $E$ episodes; add $(o_t, a^M_t)$ for every step to
        $\mathcal{D}$
      \State $\pi_k \gets \textsc{Fit}(\mathcal{D})$
    \EndFor
    \State $\hat{\pi} \gets$ the student among $\pi_0, \dots, \pi_K$ that performs best on validation episodes
    \State Initialise the PPO policy $\pi_\theta$ $\hat{\pi}$ (value function untrained); train it with PPO
      for $T$ steps in $\mathcal{E}$
    \State \Return $\pi_\theta$
  \end{algorithmic}
\end{algorithm}

\subsection{Describing the Task}
\label{sec:querying}
The description function $\phi$ is the only part of DMGT that is written for a particular environment, and it is only written once. At every step, it builds a query of the form described in \cref{sec:background-decision-models}. In natural language, the description concisely explains the environment, what the agent can control, what the goal is and the state's current values, and where useful, the last few steps with the action taken at each step. It can also include a short strategy hint: one or two sentences of general knowledge about the task, the kind a person would tell a new player. The query has the same layout at every step and only the state changes. The reward is never part of the query, the DM only acts based on the goal provided. The action of the option it chooses is $a^M_t$, and whenever the DM acts, this action is applied in the environment.

\subsection{Transferring by Imitation}
\label{sec:learning-from-dm}
The DM cannot learn from repeated interactions with the environment and the rewards, and PPO cannot act from a query; it needs a policy network that maps the observation $o_t$ to an action. We therefore train such a network, the student, to imitate the DM. The student only sees the $o_t$ and it has the same architecture as PPO's policy, so that it can directly initialise the PPO policy. Behaviour cloning fits the student to the pairs $(o_t, a^M_t)$ from $N$ episodes in which the DM acts. Because a cloned student can drift into states that the DM never visited, DAgger then lets the student act and asks the DM to label every state it reaches, building each query from the student's own history. A DM suits DAgger algorithm well, since DAgger needs an expert that can be asked for its action in any state, and a DM can provide this by answering one more query on demand, without a human in the loop.

\subsection{Improving from Reward}
\label{sec:finetuning}
PPO's policy is initialised with the weights of the selected student $\hat{\pi}$. Imitation teaches actions but not values, so the value function is still untrained, same as in PPO from scratch. PPO is then trained with default hyperparameters, to ensure consistency between DMGT and PPO from scratch, with the only difference being the policy's starting weights. PPO from scratch begins with random behaviour, randomly exploring the environment and it must find a good behaviour by this random exploration. Whereas, with DMGT the behaviour learned from the DM guides RL model to explore potentially higher value states from the start. We can therefore expect it to reach higher returns in fewer environment steps, which \cref{sec:results} tests. This is also where the blindfold comes off: the agent learns from the environment reward for the first time and can improve beyond the DM. DMGT does not rely on PPO in particular, the student can initialise the policy of any RL algorithm that uses a policy network, although we evaluate only PPO.

\begin{figure}[t]
  \centering
  \includegraphics[width=0.9\linewidth]{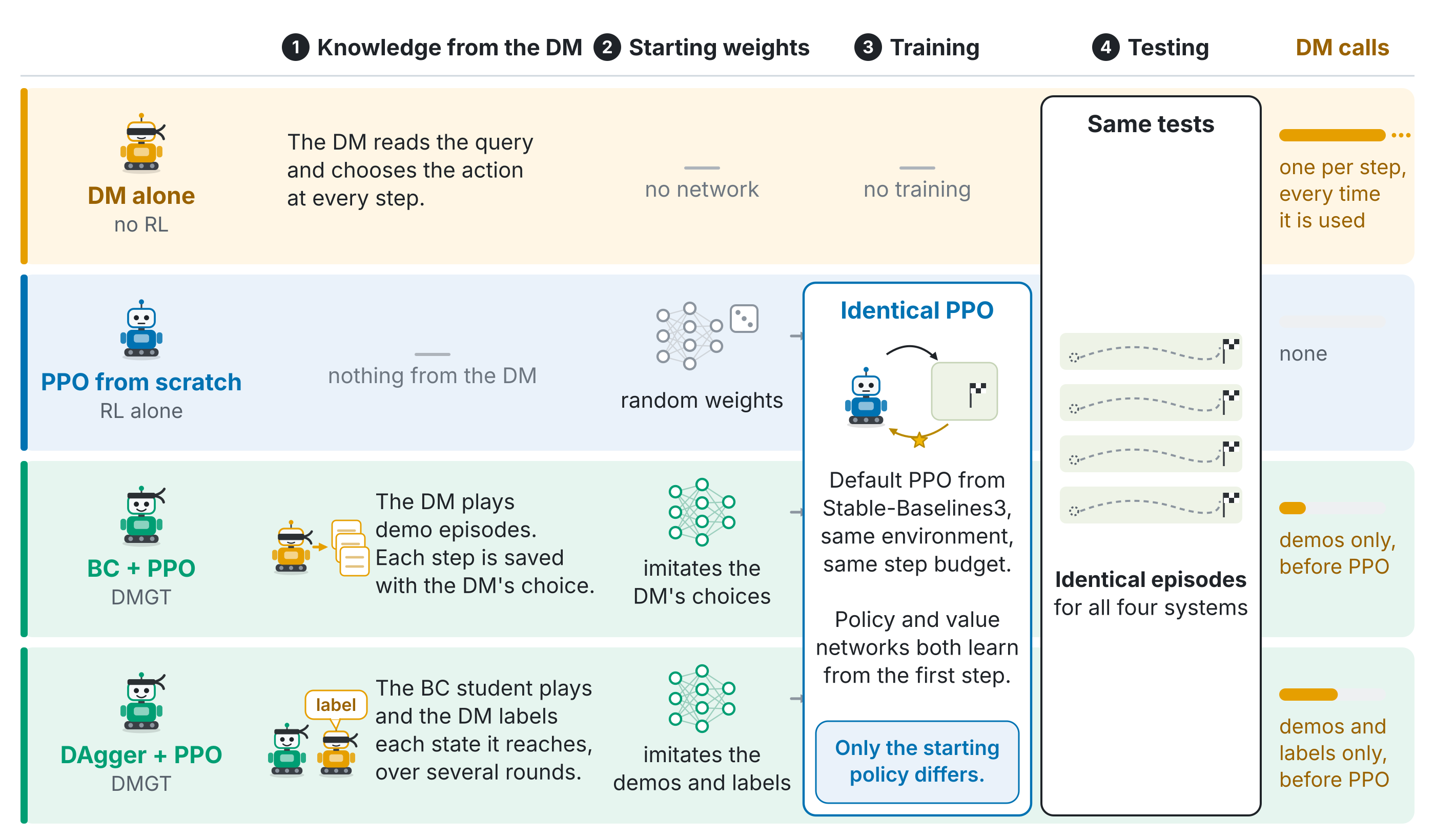}
  \caption{Rows are the four systems compared and columns are
    their stages. \emph{DM alone}: the decision model (DM) reads a text query
    describing the environment and the current state, and chooses every action, with
    no network and no training. \emph{PPO from scratch} is reinforcement learning (RL) alone, starting from random weights. The two Decision Model-Guided Training (DMGT) systems first train the policy network to imitate the DM, from the environment's observation alone. In \emph{DMGT (BC)} (behaviour cloning), the DM plays demonstration (demo) episodes and the network learns the DM's choices. In \emph{DMGT (DAgger)}, this BC student then plays, the DM labels the states it reaches, and the network learns from the demos and labels, over several rounds. All three PPO runs are otherwise identical, so only the starting policy differs. All four systems are tested on the same held-out episodes, each taking its most likely action. The bars show the DM calls each system needs. The DM alone needs one at every step for as long as it is used, while DMGT needs them only before PPO.}
  \label{fig:design}
\end{figure}

\section{Experimental Setup}
\label{sec:setup}
We evaluate Decision Model-Guided Training (DMGT) on seven tasks in six environments and compare it with PPO from scratch and also with the Decision Model (DM). Along with the final performance, we also measure the number of environment steps PPO and DMGT need to reach a given level. Additionally, an ablation study of the query breaks down the parts of the description that the DM needs, and a guideline is formed (\cref{sec:finding-what-dm-needs}).

\subsection{Environments}
\label{sec:environments}
We use six environments from Gymnasium \citep{towersGymnasiumStandardInterface2024}, highway-env \citep{leurentEnvironmentAutonomousDriving2018}, MiniGrid \citep{chevalier-boisvertMinigridMiniworldModular2023} and the Arcade Learning Environment \citep{bellemareArcadeLearningEnvironment2013, machadoRevisitingArcadeLearning2018}, with MultiRoom in two sizes (N4-S5 and N6), and together they cover different types of action space, observation space and rewards. We chose them because RL alone struggles in them: the reward is sparse or delayed, or reaching it takes a long and precise sequence of actions, so PPO alone learns slowly or not at all. The exception is Highway-fast, where PPO learns quickly; it shows how DMGT does when RL alone already works. Every environment is used as \texttt{gym.make} returns it, with its own reward.

\begin{figure}[t]
  \centering
  \setlength{\fboxsep}{0pt}
  \setlength{\fboxrule}{0.4pt}
  \newcommand{\envpanel}[3]{\begin{minipage}[t]{0.155\linewidth}\centering\fcolorbox{black!25}{white}{\includegraphics[width=\dimexpr\linewidth-0.8pt\relax]{figures/#1.png}}\\[1pt]{\footnotesize #2\par}{\scriptsize\ttfamily #3\par}\end{minipage}}
  \envpanel{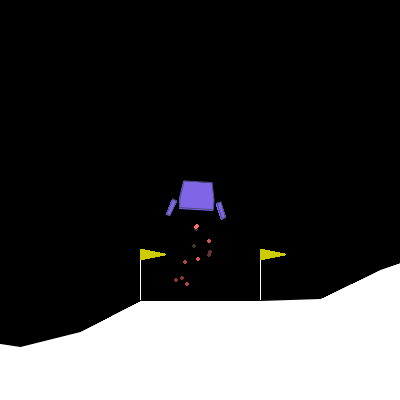}{Lunar Lander}{LunarLander-v3}\hfill
  \envpanel{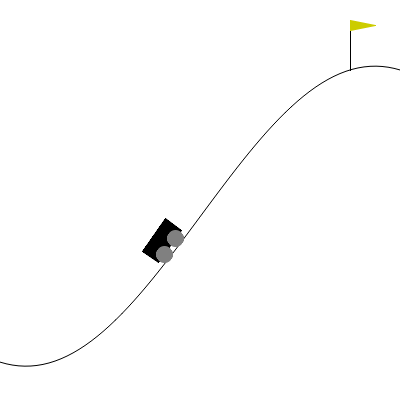}{Mountain Car}{MountainCar-\\Continuous-v0}\hfill
  \envpanel{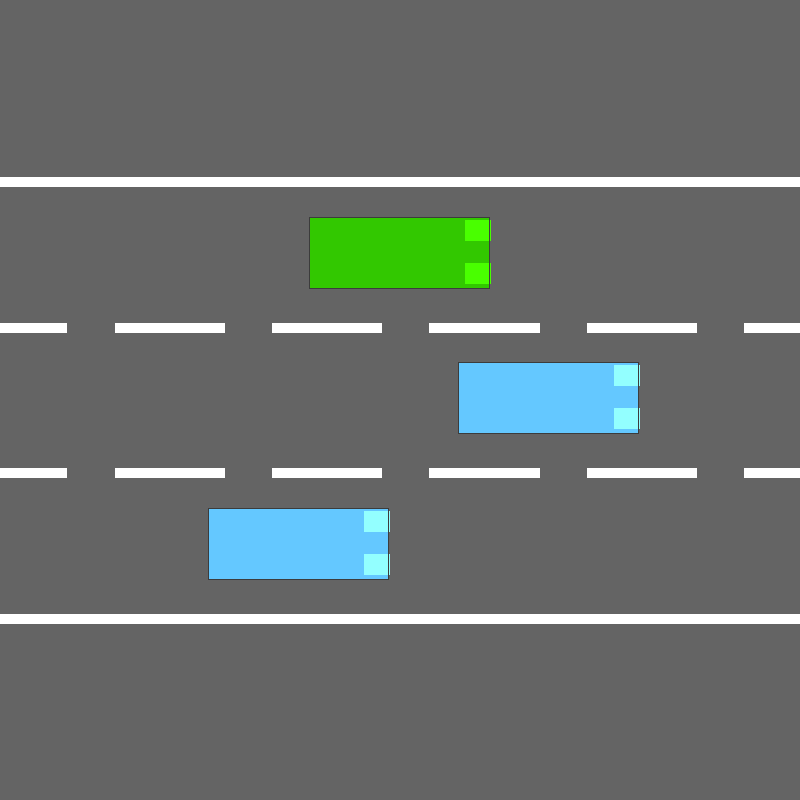}{Highway-fast}{highway-fast-v0}\hfill
  \envpanel{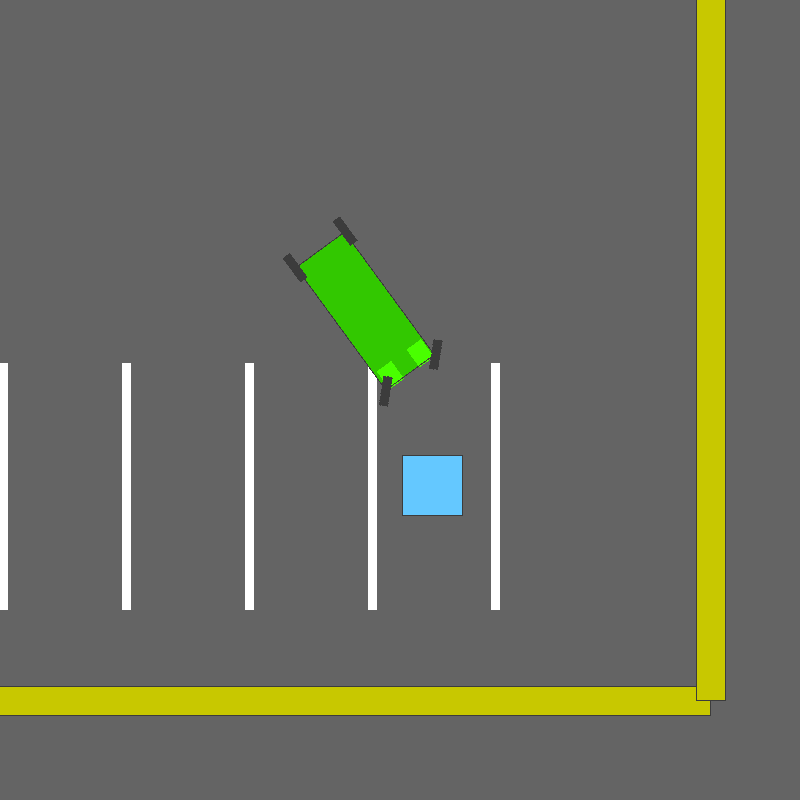}{Parking}{parking-v0}\hfill
  \envpanel{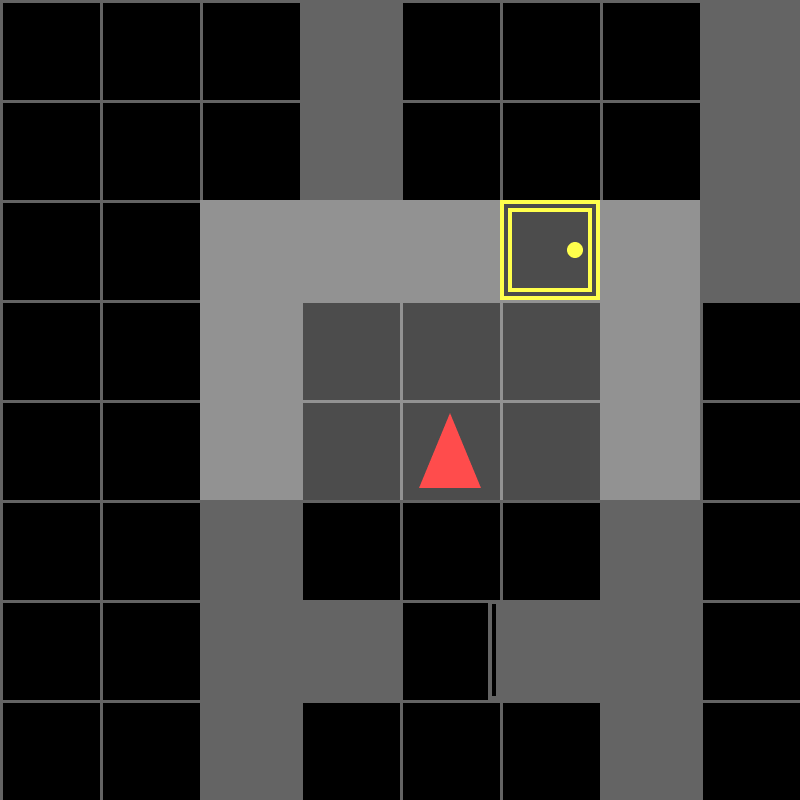}{MultiRoom N4-S5}{MiniGrid-MultiRoom-\\N4-S5-v1}\hfill
  \envpanel{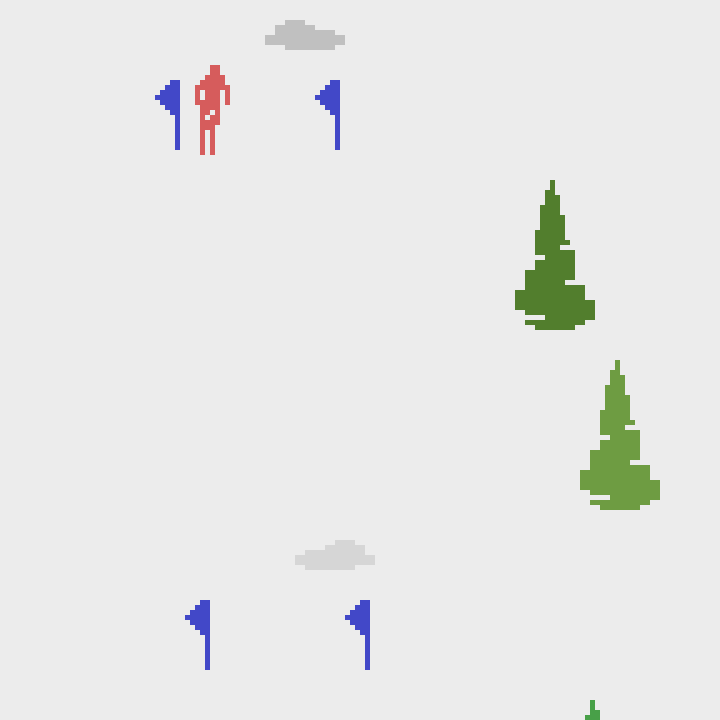}{Skiing}{ALE/Skiing-v5}
  \caption{The six environments, each shown at one step of an episode played by the decision model (DM). MultiRoom is used in two sizes, N4-S5 (shown) and N6. \Cref{app:environments} describes each one.}
  \label{fig:environments}
\end{figure}

\subsection{Decision Models}
\label{sec:setup-decision-models}
We use d3-turbo \citep{vllmsemanticrouterteamD3turboMultimodalDecision2026}, a pretrained DM that reads text and returns a probability for every answer option. It is never trained or fine-tuned on our environments and never sees their reward or a demonstration, so every decision it makes is zero-shot, from the query alone. In each environment, d3-turbo is both the DM alone (baseline) and the DM that DMGT learns from.

\subsection{Methods Compared}
\label{sec:baselines}
We compare four methods (\cref{fig:design}): First, \textbf{PPO from scratch} starts from a randomly initialised policy and learns from the environment reward alone. Second, the \textbf{DM alone} chooses every action from the query described in \cref{sec:querying}; it never sees the reward and does not learn. Third,  \textbf{DMGT (BC)} runs \cref{alg:method} without DAgger ($K = 0$): the policy is initialised by behaviour cloning on the DM's demonstration episodes and then trained with PPO. Lastly, \textbf{DMGT (DAgger)} adds DAgger iterations ($K > 0$) before PPO. The two DMGT variants and PPO from scratch use the same network, hyperparameters and number of environment steps, and differ only in the policy's starting weights. PPO from scratch thus shows what the reward alone achieves, and the DM alone what the description alone achieves.

\subsection{Training and Evaluation Protocol}
\label{sec:protocol}
\textbf{Training.} All PPO runs use the Stable-Baselines3 implementation \citep{raffinStableBaselines3ReliableReinforcement2021} with its default hyperparameters and network. Within an environment, every PPO run trains for the same number of steps: 200k in Mountain Car, 500k in Highway-fast, 1M in Lunar Lander and Parking, 2M in \mbox{MultiRoom N4-S5} and \mbox{MultiRoom N6}, and 5M in Skiing. Each PPO run uses five training seeds.

\textbf{Evaluation.} All four methods are tested on the same fixed episodes. Their seeds are not used for demonstrations, DAgger or validation, and trained policies take their most likely action. The main measure is the share of episodes that land, park or reach the goal in Lunar Lander, Parking and \mbox{MultiRoom N4-S5}, the return in Mountain Car and Highway-fast, and the final time on the game clock in Skiing, with 5\,s added for every missed gate. For PPO from scratch and both DMGT methods, each of the five trained policies plays all the test episodes, and we report the mean and standard deviation over all these episodes; the DM alone plays each test episode once. We also report learning curves of the return and of success during training, such as landing, parking or reaching the goal. To measure learning speed, we report the number of environment steps PPO needs to reach a set return, such as 200 in Lunar Lander.

\subsection{Query Ablations}
\label{sec:finding-what-dm-needs}
To find which parts of a query a DM needs, we take the full query of each environment apart, using the DM alone, without RL. Each configuration changes the full query, and the DM plays the same 3 to 20 episodes with it as with the full query. Because d3-turbo gives the same answer to the same query, one run per episode is enough, and any difference from the full query comes from the change. We first remove each part of the query on its own, such as a sentence of the description, a state value or the recent history. We also remove groups of parts together and change how some parts are written, for example giving a value as a number instead of in words. We then search for the smallest query: parts are removed, and each removal is kept only if the DM still does as well as with the full query, within a tolerance set for each environment. Because the result of this search depends on the order in which parts are removed, in most environments we check it by removing each remaining part, and pairs of parts, once more. We then test the smallest query against the full query on more episodes. From these results, and from the versions of each query that came before the ablations, we derive a detailed guideline on how to query to a DM(\cref{sec:rq3}).
\section{Results}
\label{sec:results}

\subsection{DMGT against PPO and DM}
\label{sec:rq1}
\Cref{tab:mean-return} gives the return of the four methods on the test episodes. DMGT has the highest return in five out of seven tasks, both DMGT methods are ahead of PPO from scratch and of the DM alone in Lunar Lander, Mountain Car and Skiing, DMGT (DAgger) on Highway-fast and DMGT (BC) in Parking. In Lunar Lander, the return of DMGT also varies far less from one test episode to the next than that of either. In every task, at least one DMGT method consistently outperforms PPO from scratch, and only in the two MultiRoom environments does the DM alone keep the highest return.

PPO from scratch fails in four of the tasks (\cref{fig:learning-success}): it never reaches the flag in Mountain Car, never parks in Parking and never reaches the goal in either MultiRoom environment. Instead, it settles on a behaviour that avoids the task: in Mountain Car it learns to stand still, so that its return rises to 0 and stays there (\cref{fig:learning-curves}), and in Parking it hits a wall in every episode. DMGT succeeds in all four from its first checkpoint and keeps succeeding throughout training: it reaches the flag in every evaluation episode, the goal in about 98\% of them in \mbox{MultiRoom N4-S5} and in about three quarters in \mbox{MultiRoom N6}, and parks in about three quarters. In Skiing, DMGT misses almost no gates during training, whereas PPO from scratch still misses about one per episode at the end.

\begin{table}[t]
  \setlength{\belowcaptionskip}{6pt}
  \caption{Mean return on the test episodes, $\pm$ the standard deviation over these episodes, pooled over the five training seeds for PPO from scratch and both DMGT methods. The return is the sum of the rewards that the environment gives over one episode, so higher is better in every row. The best mean in each row is in bold.}
  \label{tab:mean-return}
  \centering
  \small
  \begin{tabular}{@{}lrrrr@{}}
    \toprule
    & PPO from & DM & DMGT & DMGT \\
    Environment & scratch & alone & (BC) & (DAgger) \\
    \midrule
    \texttt{LunarLander-v3} & 185.3 $\pm$ 86.1 & 209.0 $\pm$ 125.8 & 263.5 $\pm$ 26.6 & \textbf{269.3 $\pm$ 18.1} \\
    \texttt{MountainCarContinuous-v0} & 0.0 $\pm$ 0.0 & 88.0 $\pm$ 1.0 & \textbf{93.3 $\pm$ 0.7} & 93.2 $\pm$ 0.6 \\
    \texttt{highway-fast-v0} & 21.7 $\pm$ 0.4 & 22.7 $\pm$ 2.2 & 21.5 $\pm$ 4.2 & \textbf{22.9 $\pm$ 0.5} \\
    \texttt{parking-v0} & $-$16.9 $\pm$ 0.9 & $-$19.4 $\pm$ 9.3 & \textbf{$-$15.6 $\pm$ 5.7} & $-$26.2 $\pm$ 40.7 \\
    \texttt{MiniGrid-MultiRoom-N4-S5-v1} & 0.000 $\pm$ 0.000 & \textbf{0.763 $\pm$ 0.084} & 0.679 $\pm$ 0.253 & 0.727 $\pm$ 0.186 \\
    \texttt{MiniGrid-MultiRoom-N6-v0} & 0.000 $\pm$ 0.000 & \textbf{0.604 $\pm$ 0.127} & 0.471 $\pm$ 0.291 & 0.360 $\pm$ 0.328 \\
    \texttt{ALE/Skiing-v5} & $-$3,899 $\pm$ 701 & $-$3,677 $\pm$ 16 & \textbf{$-$3,348 $\pm$ 60} & $-$3,351 $\pm$ 48 \\
    \bottomrule
  \end{tabular}
\end{table}

\subsection{Learning Speed}
\label{sec:rq2}
\begin{figure}[t]
  \centering
  \includegraphics[width=\linewidth]{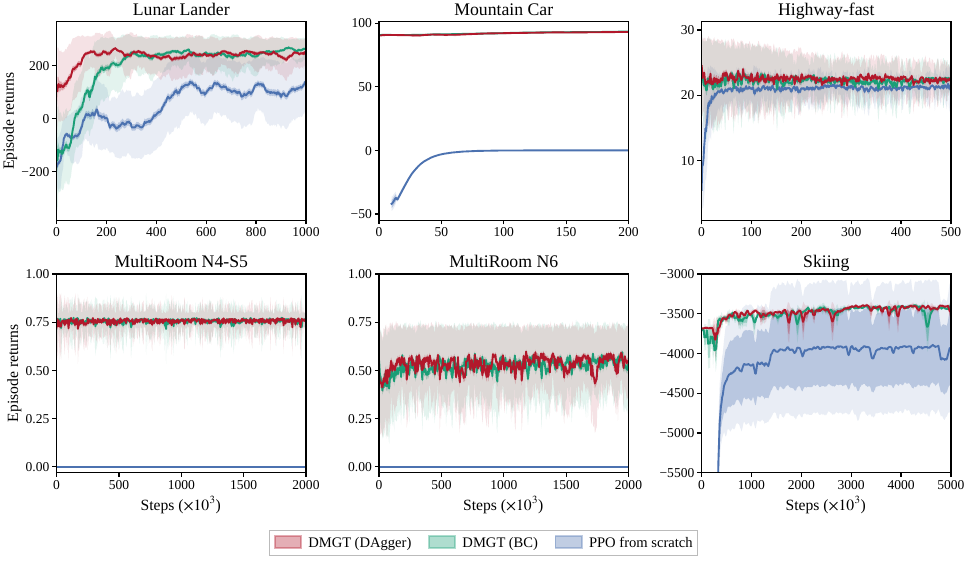}
  \caption{Return during training for PPO from scratch, DMGT (BC) and DMGT (DAgger), one panel per task. The x-axis counts PPO's environment steps. Each line is the mean return over the last 100 training episodes, or over the episodes in the last 5\% of the steps shown when these are fewer. The dark band is the standard error and the light band one standard deviation either side.}
  \label{fig:learning-curves}
\end{figure}

\begin{figure}[t]
  \centering
  \includegraphics[width=\linewidth]{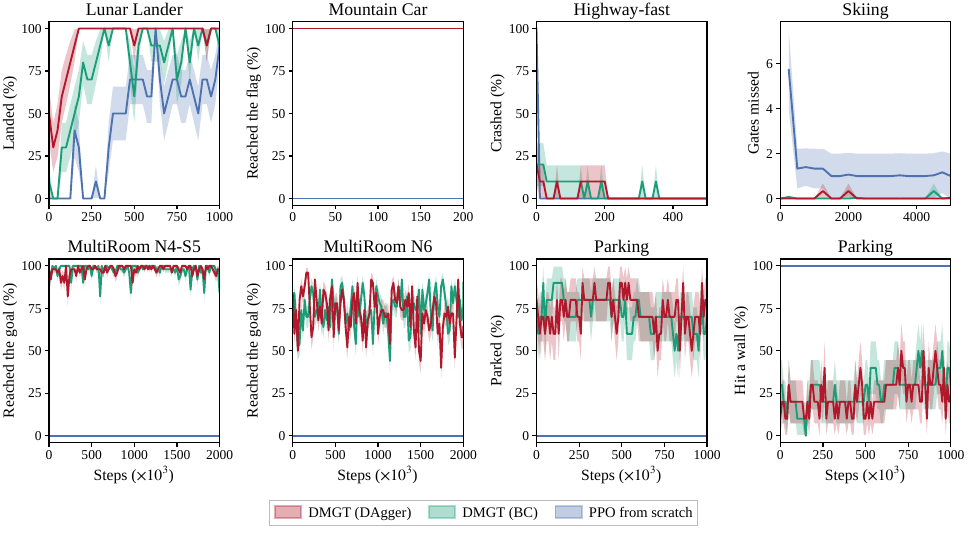}
  \caption{Success during training for PPO from scratch, DMGT (BC) and DMGT (DAgger), in the environments of \cref{fig:learning-curves} and in Parking. At checkpoints during training (every 10k steps; 20k in MultiRoom, 25k in Lunar Lander and 250k in Skiing), the policy plays fixed evaluation episodes, other than the test episodes, with its most likely action: 10 per checkpoint, 5 in Mountain Car and 50 in MultiRoom. Each panel gives the share of these episodes in which the agent lands, reaches the flag, crashes, reaches the goal, parks or hits a wall, or, in Skiing, the gates missed per episode that reaches the finish; lower is better for crashed, gates missed and hit a wall. The band is the standard error over the five training seeds. }
  \label{fig:learning-success}
\end{figure}

DMGT starts from the DM's behaviour, so it begins training at a level that PPO from scratch reaches late or never (\cref{fig:learning-curves}). DMGT (DAgger) stays above PPO from scratch throughout training in every environment, and DMGT (BC) does too, except between 21k and 59k steps in Lunar Lander and in brief dips on Highway-fast. In Mountain Car and both MultiRoom environments, the return of DMGT is close to its final level from the first training episodes. On Highway-fast, the first 100 training episodes of DMGT already average a higher return than PPO from scratch reaches at the end of training (22.5 with BC and 23.1 with DAgger, against 21.3), while PPO from scratch needs 26.5k steps to come within 5\% of its own final level. In Skiing, PPO from scratch starts far below DMGT and stays below it for the whole 5M steps. In Lunar Lander, the average return of DMGT (DAgger) reaches 200 after 83k steps and that of DMGT (BC) after 216k, a level that PPO from scratch does not reach in 1M steps. The landings on the evaluation episodes follow the same order (\cref{fig:learning-success}): DMGT (DAgger) first lands in all of them after 175k steps, DMGT (BC) after 325k and PPO from scratch after 625k, and over the whole of training they land in 92.5\%, 77\% and 42.5\% of them.

\subsection{Guidelines on how to Query a Decision Model}
\label{sec:rq3}
\Cref{fig:guidelines} the results of the query ablations as guidelines for describing an environment and task to a DM. They follow from one view of the DM, it is like a blindfolded capable player, but the player has never played the game before and forgets everything between turns, so its only context is what the query tells it. A query therefore has two parts, a briefing, which is the same at every step, and a description of the current state, which keeps its form while its numbers change from one step to the next. A well-designed query lets the DM perform well. In \mbox{MultiRoom N4-S5}, with a query that describes the world, along with the observational data and the observational data's explanation, and some historic data, the agent reached the goal in all test episodes.

The query is what lets the DM act. Given only the raw observations that PPO receives in place of the description, the DM does not reach the goal in any of the test episodes of \mbox{MultiRoom N4-S5} and parks in none of the test episodes of \mbox{Parking}. 

An accurate description of each action is essential. In Mountain Car, with only the names of the actions, the DM pushes right on every step and never reaches the flag, and when nothing says what the actions do, it never parks in Parking. The facts the DM cannot guess matter together, not one by one: on Highway-fast, removing any one of them barely changes the return, but removing all of them causes a collision. Rewording the goal, on the other hand, changes almost nothing.

In the description of the current state, positions are best given from the agent's point of view: in Parking, a shorter query parks in some test episodes from the driver's seat, but in none in the car park's coordinates. What is directly in front of the agent also has to be described, and each value needs to be given only once, as a number or in words. Some history helps, but more does not, and it keeps each query different from the last, which matters because the DM gives the same answer to the same query: in many of the loops in which the DM got stuck in \mbox{MultiRoom N4-S5}, the query did not change from one step to the next.

The surface of the query, such as key names, the order of options and fields and the number of decimals, makes no measurable difference, and a shorter query can work as well: in Mountain Car, less than half of the tokens still reach the flag in every episode. Because any change can send an episode down a different path, a query has to be tested on several episodes, and a shortened one on more episodes than it was found on.

These results give the guidelines in \cref{fig:guidelines}; the ablation results behind each one are in \cref{app:guidelines}.

\begin{figure}[t]
\centering
\fcolorbox{black!30}{black!3}{%
\begin{minipage}{\dimexpr\linewidth-2\fboxsep-2\fboxrule\relax}
\small
\vspace{2pt}
\textit{The environment and the task, the same at every step}
\begin{enumerate}
  \setlength{\itemsep}{1pt}\setlength{\parsep}{0pt}\setlength{\topsep}{2pt}
  \item \textbf{The goal.} State the goal as the environment scores it, and nothing else. Rewording the goal will not steer behaviour.
  \item \textbf{The options.} Make each option a decision that matters, and leave low-level control, such as keeping a steady speed, to code. Then check that the task can still be done when a choice is made only as often as the DM is asked.
  \item \textbf{The actions.} Describe each action accurately; this is essential. Give each option a name that says what it does and a description of its measured effect, including when it does nothing.
  \item \textbf{One place for each action.} Explain what each action does in one place only: its option. Do not describe the actions again elsewhere in the query.
  \item \textbf{The facts.} Give the facts the DM cannot guess, measured in the environment and with units, together in one place: how long each choice lasts, what ends the episode and how it is scored, what the agent can see, and anything surprising.
\end{enumerate}

\textit{The moment, whose numbers change at every step}
\begin{enumerate}
  \setlength{\itemsep}{1pt}\setlength{\parsep}{0pt}\setlength{\topsep}{2pt}
  \setcounter{enumi}{5}
  \item \textbf{The point of view.} Describe the moment from the agent's point of view; map coordinates also work with context. Say what is directly in front of the agent, define left and right once, state what each sign means, and use metres, seconds and degrees, never pixels.
  \item \textbf{No repetition.} Do not repeat information, especially in a different format. Repetition can be harmful, and removing it helps slightly.
  \item \textbf{Some history.} Give the step number, the steps left, and the last few choices with what each did, but not much more. History also keeps each query different: given an identical query, a DM may repeat a move that changed nothing, such as walking into a wall.
  \item \textbf{Words, not pictures.} Describe; never send a picture or the raw readings.
\end{enumerate}

\textit{Before relying on a query}
\begin{enumerate}
  \setlength{\itemsep}{1pt}\setlength{\parsep}{0pt}\setlength{\topsep}{2pt}
  \setcounter{enumi}{9}
  \item \textbf{The surface.} Keep the description of the environment and the task in the state, not the question. Do not polish the surface: a shorter query often works as well.
  \item \textbf{Testing.} Run several episodes, including hard ones, check any shortened query on new episodes, and log every query, every answer and which DM answered.
\end{enumerate}
\vspace{1pt}
\end{minipage}}
\caption{Guidelines for querying a decision model, drawn from the query ablations.}
\label{fig:guidelines}
\end{figure}

\section{Conclusion}
\label{sec:conclusion}

In this paper, we introduced Decision Model-Guided Training (DMGT), which combines the complementary strengths of a decision model and reinforcement learning. A DM acts well from a description of the environment and the task in natural language, without any training, but it cannot learn from reward, while RL improves from reward but starts from scratch. DMGT uses the DM's decisions to initialise an RL policy through imitation learning, and RL then improves this policy from the reward. Across seven tasks in six environments, DMGT outperforms RL alone in all seven tasks and the DM alone in five, and it starts training at a level that RL alone reaches late or never. Where RL alone settles on a behaviour that avoids the task, such as standing still in Mountain Car or hitting a wall in Parking, DMGT succeeds from the start of training. Once trained, the policy acts without the DM, so the DM is only needed before training.

Through query ablations, we also showed that what a DM is told decides whether it succeeds. The DM cannot act on the raw observations that RL receives, an accurate description of each action is essential, and surface details of the query make no measurable difference. We collected these results as guidelines for describing an environment and task to a DM, which treat it as a blindfolded player who knows only what the query tells it. Together, these results show that a DM can give RL a head start without behaviour that must first be recorded or trained for the task: it only needs a good description of the environment and the task.

\FloatBarrier

\bibliography{references}
\bibliographystyle{tmlr}

\appendix
\section{Decision Model Queries}
\label{app:requests}
In every environment, the DM receives one query per decision, with the three parts of a query: a description of the situation, written as a JSON object; a question, which asks what the agent does next; and the options, which are the actions, each with a name and a description of what it does. The description opens with a briefing, the same at every step, that describes the world, the scoring, the measured facts the DM cannot guess, a short strategy hint and how to read the values. The rest of the description, the moment, describes the current state and changes at every step. Below, for each environment, are the goal, the options, what the moment contains, and the size and rate of the queries. The full queries are in the code (\cref{app:reproducibility}).

\paragraph{Lunar Lander.} Goal: ``land the lander. Bring it down to the ground so that it comes to rest on its legs, without its body touching the ground.'' Options: \texttt{do\_nothing}, \texttt{fire\_left\_engine}, \texttt{fire\_main\_engine} and \texttt{fire\_right\_engine}, each described by what the engine does to the lander. The moment: the lander's position and height above the pad, its sideways and vertical speed, its tilt and turning, whether each leg touches the ground, where the pad is, and the last 10 steps. About 2230 tokens per query, one query per environment step (50 per second), 348 per test episode.

\paragraph{Mountain Car.} Goal: ``get the car to the flag.'' Options: \texttt{push\_left}, \texttt{no\_push} and \texttt{push\_right}: full force to the left, no force, or full force to the right. The moment: the car's position, its height above the valley floor and its speed, where the flag is, the last 10 steps, and the episode so far (the step, the steps left and the furthest positions reached). About 1290 tokens per query, one query per step, 120 per test episode.

\paragraph{Highway-fast.} Goal: ``score as much as possible over the 30-second drive: drive as fast as you can without colliding with another car, in the right lane when you can.'' Options: \texttt{change\_lane\_left}, \texttt{keep\_going}, \texttt{change\_lane\_right}, \texttt{speed\_up} and \texttt{slow\_down}, each with its measured effect and when it does nothing. The moment: the time; the car's lane and speed; for its own lane and each lane beside it, the car ahead, its gap and speed, the closing speed and the seconds until the car would reach it; the cars in view; and the last 5 choices. About 2250 tokens per query, one query per second, 29 per test episode.

\paragraph{Parking.} Goal: ``park the car in your space, as quickly as you can, without hitting a wall.'' Options: \texttt{forward\_left}, \texttt{forward\_straight}, \texttt{forward\_right}, \texttt{back\_up\_left}, \texttt{back\_up\_straight}, \texttt{back\_up\_right} and \texttt{stop}: drive forward or backward with the wheels turned fully left, straight or fully right, or stop. The moment: the time, the car's speed, where the parking space is from the driver's seat and which way it points, how far the walls are ahead and behind, and the last 5 choices. About 1800 tokens per query, five queries per second, 54 per test episode.

\paragraph{MultiRoom N4-S5 and N6.} Goal: ``reach the green goal square in as few steps as possible'', followed by how this is scored, with 80 steps in \mbox{MultiRoom N4-S5} and 120 in \mbox{MultiRoom N6}. Options: MiniGrid's seven actions (\texttt{turn\_left}, \texttt{turn\_right}, \texttt{move\_forward}, \texttt{pick\_up}, \texttt{drop}, \texttt{toggle} and \texttt{done}), each described with what it does here, including when it does nothing. The moment: the time; the agent's view as a text map; what is right in front of it; the doors and the goal it has seen, where they are and how many steps away; the doors it has opened; and the last 5 choices with what each did. About 1730 tokens per query in N4-S5 and 1820 in N6, one query per step, 21 and 52 per test episode.

\paragraph{Skiing.} Goal: ``ski through all 20 gates, between the two flags of each, and reach the finish as fast as you can.'' Options: \texttt{turn\_left}, \texttt{keep\_direction} and \texttt{turn\_right}: turn the skis one notch towards the left or the right of the screen, or keep them as they are. The moment: the time; the direction of the skis and the skier's downhill and sideways speed; the gates reached, passed and missed; where the next gate is and how soon the skier reaches it; up to three trees ahead; and the last 5 choices. About 1820 tokens per query, 15 queries per second, about 550 per test episode.

\section{Environment Details}
\label{app:environments}
\Cref{tab:environments} describes each environment as PPO sees it and as the DM acts in it.

The test episodes use the environment seeds from 10000 (10000--10019; 10000--10009 in Mountain Car; 10000--10099 in MultiRoom), the DAgger episodes the seeds from 20000, the evaluation episodes during training those from 30000, and the validation episodes those from 40000. The DM's demonstrations use the seeds 0--99, 0--19 in Skiing, 0--7 in Mountain Car and 0 in Lunar Lander.

\begin{table}[ht]
  \caption{The environments: what PPO observes, the actions of PPO and of the DM, the reward and the length of an episode.}
  \label{tab:environments}
  \centering
  \footnotesize
  \begin{tabular}{@{}p{0.12\linewidth}p{0.18\linewidth}p{0.22\linewidth}p{0.27\linewidth}p{0.09\linewidth}@{}}
    \toprule
    Environment & Observation & Actions & Reward & Episode \\
    \midrule
    Lunar Lander & 8 values: position, velocity, tilt, turning rate and the contact of each leg & 4: no engine, or the left, main or right engine; the DM chooses among the same four & shaped towards coming to rest on the pad: $+100$ for coming to rest, $-100$ for a crash, $+10$ for each leg on the ground, small costs for firing the engines & 1000 steps, 50 per second \\
    \addlinespace
    Mountain Car & 2 values: position and velocity & a force in $[-1, 1]$; the DM's three options apply $-1$, 0 or $+1$ & $+100$ at the flag, $-0.1\,a^2$ per step for the force $a$ & 999 steps \\
    \addlinespace
    Highway-fast & $5 \times 5$: presence, position and velocity of the car and of the 4 nearest cars & 5: change lane left, keep going, change lane right, faster, slower; the DM chooses among the same five & up to 1 per second for speed between 20 and 30\,m/s and for the right lane; a collision ends the drive & 30 decisions, one per second \\
    \addlinespace
    Parking & 18 values: position, velocity and heading of the car and of its space & acceleration and steering in $[-1, 1]$; the DM's seven options (forward or backward with the wheels left, straight or right, or stop) set these controls & minus the square root of the weighted distance to the space at every step, and $-5$ for hitting a wall; parked when the reward is above $-0.12$ & 100\,s, five decisions per second \\
    \addlinespace
    MultiRoom N4-S5 and N6 & the $7 \times 7 \times 3$ image of the agent's view & MiniGrid's 7 actions; the DM chooses among the same seven & $1 - 0.9\,t/T$ on reaching the goal after $t$ steps, 0 otherwise & $T = 80$ or 120 steps \\
    \addlinespace
    Skiing & object features read off the screen & 3: left, none, right; the DM chooses among the same three & minus the time on the game clock, with 5\,s added for each of the 20 gates missed & until the finish, or 5 minutes on the clock \\
    \bottomrule
  \end{tabular}
\end{table}

\section{Training Details and Hyperparameters}
\label{app:training}
PPO uses Stable-Baselines3 2.9.0; \cref{tab:hyperparameters} lists its default hyperparameters and networks. DMGT starts PPO from the student's policy network, and in Mountain Car and Parking also from its action spread.

\Cref{tab:imitation} gives the imitation in each environment. Behaviour cloning minimises the cross-entropy of the DM's choice, or, for the continuous actions of Mountain Car and Parking, the negative log-likelihood of the action that the choice applies. It uses Adam with a learning rate of $10^{-3}$ and batches of 64, for 200 epochs in Lunar Lander and Mountain Car and for at most 15000 updates elsewhere (60000 in Parking). In Mountain Car, the inputs are scaled by the demonstrations' mean and standard deviation, and this scaling is folded into the first layer, so that PPO receives the raw observation. Each DAgger iteration has the student play its episodes, labels every state it reaches with the DM's choice, and trains the student again on all the data.

\begin{table}[ht]
  \caption{PPO hyperparameters: the defaults of Stable-Baselines3 2.9.0.}
  \label{tab:hyperparameters}
  \centering
  \footnotesize
  \begin{tabular}{@{}ll@{}}
    \toprule
    Hyperparameter & Value \\
    \midrule
    Policy and value networks & separate, two hidden layers of 64 units, tanh \\
    Environments $\times$ steps per update & $1 \times 2048$ \\
    Minibatch size & 64 \\
    Epochs per update & 10 \\
    Learning rate & $3 \times 10^{-4}$ \\
    Discount factor $\gamma$ & 0.99 \\
    GAE $\lambda$ & 0.95 \\
    Clip range & 0.2 \\
    Entropy coefficient & 0 \\
    Value-loss coefficient & 0.5 \\
    Maximum gradient norm & 0.5 \\
    Target KL divergence & none \\
    \bottomrule
  \end{tabular}
\end{table}

\begin{table}[ht]
  \caption{Imitation in each environment: the DM's demonstration episodes and their decisions, and the DAgger iterations and episodes per iteration with the DM's labels.}
  \label{tab:imitation}
  \centering
  \footnotesize
  \begin{tabular}{@{}lll@{}}
    \toprule
    Environment & Demonstrations & DAgger \\
    \midrule
    Lunar Lander & 1 (357 decisions) & $3 \times 4$ (2909 labels) \\
    Mountain Car & 8 (903) & $2 \times 2$ (454--457) \\
    Highway-fast & 100 (2828) & $5 \times 10$ (1390) \\
    Parking & 100 (4559) & $5 \times 10$ (3140) \\
    \mbox{MultiRoom N4-S5} & 100 (2075) & $5 \times 50$ (5360) \\
    \mbox{MultiRoom N6} & 100 (5292) & $5 \times 50$ (19802) \\
    Skiing & 20 (10880) & $5 \times 4$ (11103--11115) \\
    \bottomrule
  \end{tabular}
\end{table}

\section{Additional Results}
\label{app:guidelines}
Table~\ref{tab:guideline-evidence} gives, for each guideline in Figure~\ref{fig:guidelines}, the ablation results it is based on. Each result compares a changed query with the full query on the same episodes, with the DM alone and no RL. The number of episodes is given with each result.

\begin{table}[ht]
  \caption{The ablation results behind each guideline of \cref{sec:rq3}.}
  \label{tab:guideline-evidence}
  \centering
  \footnotesize
  \begin{tabular}{@{}p{0.22\linewidth}p{0.74\linewidth}@{}}
    \toprule
    Guideline & Evidence \\
    \midrule
    1.\ The goal & Highway-fast: three rewordings of the goal change the return by at most 0.1. \\
    \addlinespace
    2.\ Accurate action descriptions & Mountain Car: names without descriptions, no run reaches the flag (0 of 2), pushing right on all 300 steps. Neutral names without descriptions: no episode parks in Parking (0 of 6), and \mbox{MultiRoom N4-S5} needs 23.8 more steps per episode. \\
    \addlinespace
    3.\ Actions explained once & Mountain Car: the description of the environment also said that the car can be pushed left, pushed right or not pushed. Without that sentence, so that only the options explain the actions, the car still reaches the flag (in 112 steps instead of 110), and the shortest query leaves the sentence out. \\
    \addlinespace
    4.\ Facts the DM cannot guess & Highway-fast (return out of 30): the facts matter as a group. Removed one at a time, each of the six lowers the return by at most 1.3, with no collision; removed all together, they lower it by 2.8, and the car collides in 1 of 3 episodes. A false fact, that the car brakes on its own, lowers the return by 1.1. \\
    \addlinespace
    5.\ The agent's point of view & \mbox{MultiRoom N4-S5}: with a shorter query, without the view and what is in front, 9 instead of 19 of 20 episodes reach the goal. \\
    \addlinespace
    6.\ No repetition & Highway-fast: every value only as a number or only in words changes the return by at most 0.4. Without both copies of the other cars, a shortened query collides in 2 of 3 episodes. \\
    \addlinespace
    7.\ Some history & Highway-fast: without the recent choices, 2.8 lower and a collision; with 1 to 20 of them, at most 0.1 different. Mountain Car: the shortest query fails without its one recent step. \mbox{MultiRoom N4-S5}: in 306 of 731 loops, the query repeats exactly. \\
    \addlinespace
    8.\ No pictures or raw readings & PPO's raw observation in place of the description: 0 of 100 test episodes in \mbox{MultiRoom N4-S5}, 0 of 6 in Parking, and 10.0 lower with collisions in 2 of 3 on Highway-fast; adding the description of the environment and the task does not help. Lunar Lander: with a picture at every step, the lander flies off the screen. \\
    \addlinespace
    9.\ The surface & Highway-fast: the description in the question, 1.1 lower; key names, option order, field order and decimals, at most 0.1. Mountain Car: 489 tokens instead of 1,294 reach the flag in 8 of 8 episodes. \\
    \addlinespace
    10.\ Testing & Lunar Lander: on 5 episodes, 52 of 69 changes to the query landed fewer, or different, episodes than the full query. Mountain Car: the shortest query, found on 1 episode, held on 8. \\
    \bottomrule
  \end{tabular}
\end{table}

\section{Compute and Cost}
\label{app:compute}
Training ran on a laptop (Intel Core i7-12700H) and on two servers, each with two Intel Xeon Gold 5217 CPUs and two NVIDIA Quadro RTX 8000 GPUs. The DM was queried through its hosted free API. \Cref{tab:compute} gives the time of one PPO run, the DM calls that the DM alone makes per test episode, the DM calls that DMGT makes once, for training, and the time of one call. Once trained, DMGT makes no calls. The calls include retries after failed answers.

\begin{table}[!ht]
  \caption{Compute and DM calls. PPO time: one training run, the range over the methods and seeds. DMGT calls: the demonstrations for DMGT (BC), and the demonstrations and DAgger labels together for DMGT (DAgger).}
  \label{tab:compute}
  \centering
  \footnotesize
  \begin{tabular}{@{}lrrrr@{}}
    \toprule
    & PPO time & DM calls per & DMGT calls, & Time per \\
    Environment & per run & test episode & BC; DAgger & DM call \\
    \midrule
    Lunar Lander & 18 min & 348 & 480; 5143 & 0.38\,s \\
    Mountain Car & 2--4 min & 120 & 903; 1357--1360 & -- \\
    Highway-fast & 4.1--4.2 h & 29 & 2866; 4256 & 0.61\,s \\
    Parking & 2.3--2.4 h & 54 & 5076; 8475 & 0.54\,s \\
    \mbox{MultiRoom N4-S5} & 43--44 min & 21 & 2075; 4172 & 1.50\,s \\
    \mbox{MultiRoom N6} & 44--56 min & 52 & 5292; 25094 & 1.71\,s \\
    Skiing & 1.7--1.9 h & about 550 & 16737; about 27850 & 0.48\,s \\
    \bottomrule
  \end{tabular}
\end{table}

\section{Reproducibility}
\label{app:reproducibility}
Our implementation is one Python 3.12 workspace with one folder per environment and a shared package of DM clients.  Every DM call is logged with its query, the probability of each option and the model that answered. The experiments use Gymnasium 1.4.0, highway-env 1.12.1, MiniGrid 3.2.0, ALE 0.12.1, Stable-Baselines3 2.9.0, PyTorch 2.14.1 and NumPy 2.5.4. The environment seeds are given in \cref{app:environments}.

\end{document}